\documentclass[letterpaper, 10 pt, conference]{ieeeconf}  % Comment this line out if you need a4paper

\IEEEoverridecommandlockouts                              % This command is only needed if 
\usepackage{cite}
\usepackage{amsmath,amssymb,amsfonts}
\usepackage{algorithmic}
\usepackage{graphicx}
\usepackage{textcomp}
\usepackage{xcolor}

\usepackage{bm,textcomp,booktabs,xcolor}
\usepackage{pgfplots,tikzscale}
\usepackage{amsfonts}
\usepackage{pgfgantt}
\usepackage{soul}
\usepackage{makecell}
\usepackage{amsmath, amsfonts, amssymb}
\graphicspath{{./fig/}}

\usepackage{algorithm}
\usepackage{subcaption}
\usepackage{arydshln}

\title{\LARGE \bf
Anomaly Detection Based on Selection and Weighting in Latent Space
}

\author{Yiwen~Liao, Alexander~Bartler, and~Bin~Yang% <-this % stops a space
\thanks{Yiwen Liao, Alexander Bartler and Bin Yang are with the Institute of Signal Processing and System Theory of the University of Stuttgart. Email: {\tt\small \{yiwen.liao, alexander.bartler, bin.yang\}@iss.uni-stuttgart.de}}%
}

\begin{document}

\maketitle
\thispagestyle{empty}
\pagestyle{empty}

%%%%%%%%%%%%%%%%%%%%%%%%%%%%%%%%%%%%%%%%%%%%%%%%%%%%%%%%%%%%%%%%%%%%%%%%%%%%%%%%
\begin{abstract}
With the high requirements of automation in the era of Industry 4.0, anomaly detection plays an increasingly important role in higher safety and reliability in the production and manufacturing industry. Recently, autoencoders have been widely used as a backend algorithm for anomaly detection. Different techniques have been developed to improve the anomaly detection performance of autoencoders. Nonetheless, little attention has been paid to the latent representations learned by autoencoders. In this paper, we propose a novel selection-and-weighting-based anomaly detection framework called SWAD. In particular, the learned latent representations are individually selected and weighted. Experiments on both benchmark and real-world datasets have shown the effectiveness and superiority of SWAD. On the benchmark datasets, the SWAD framework has reached comparable or even better performance than the state-of-the-art approaches.
\end{abstract}

%%%%%%%%%%%%%%%%%%%%%%%%%%%%%%%%%%%%%%%%%%%%%%%%%%%%%%%%%%%%%%%%%%%%%%%%%%%%%%%%
\section{Introduction}
\label{sec:intro}

The development of automation engineering has been significantly changing the entire industry from different perspectives. On one hand, automation technology has been nowadays widely used in diverse fields, such as autonomous driving systems~\cite{paden2016survey,gerla2014internet}, robotics~\cite{kehoe2015survey}, manufacturing~\cite{posada2015visual,susto2014machine,chen2017smart}, validation and testing~\cite{fan2020data}. The successful application of automation technology notably promotes the advancements of industrial society. On the other hand, enjoying the convenience brought by the fast advancements of automation technology, it is inevitably to raise the concerns about the reliability and safety within the technology itself. Accordingly, in order to realize a more reliable automation system, to detect and handle unexpected failures becomes an important issue in both academic and industrial communities. 

However, manual identification of failures is not feasible in the era of big data and cannot meet the efficiency requirements. Thereby, to design automatic failure analysis and detection systems is nowadays critical, especially for Industry 4.0. Fortunately, anomaly detection has been intensively studied in literature~\cite{chalapathy2019deep,wang2019progress,hodge2004survey}. Early anomaly detection methods often relied on explicit defined models for specific systems~\cite{gao2015survey}. These methods requires special expert knowledge and have limited generalization ability. Afterwards, statistical and machine learning methods drew people's attention due to their flexibility. Representative approaches include~\cite{scholkopf2001estimating,liu2008isolation,breunig2000lof}. Although these methods achieved notable advancements in anomaly detection, they typically require complex feature engineering and are not robust to different data domains.

Recently, deep learning has achieved dramatic success in many different fields, including image classification~\cite{brock2021highperformance}, natural language processing~\cite{devlin-etal-2019-bert}, as well as anomaly detection~\cite{chalapathy2019deep}. In some cases, when both normal and abnormal data are available during training, it is natural to design a neural network for binary classification and to train it in a supervised way~\cite{bartler2018automated}. However, anomalies are generally rare in practice and it is thus difficult to acquire enough abnormal samples to train neural networks. Consequently, it is more appealing to train deep-learning-based methods on normal data only~\cite{schlachter2019deep,ruff2018deep}, which are easier to collect in practice. In this context, the training itself can be considered as semi-supervised, since the model is fully trained on the normal data only~\cite{chalapathy2019deep}. In contrast to supervised anomaly detection, a limited number of abnormal samples are often used to validate the model and fine-tune the hyperparameters in the semi-supervised setup. 

\begin{figure}[!htb]
	\centering
	\includegraphics[width=.95\linewidth]{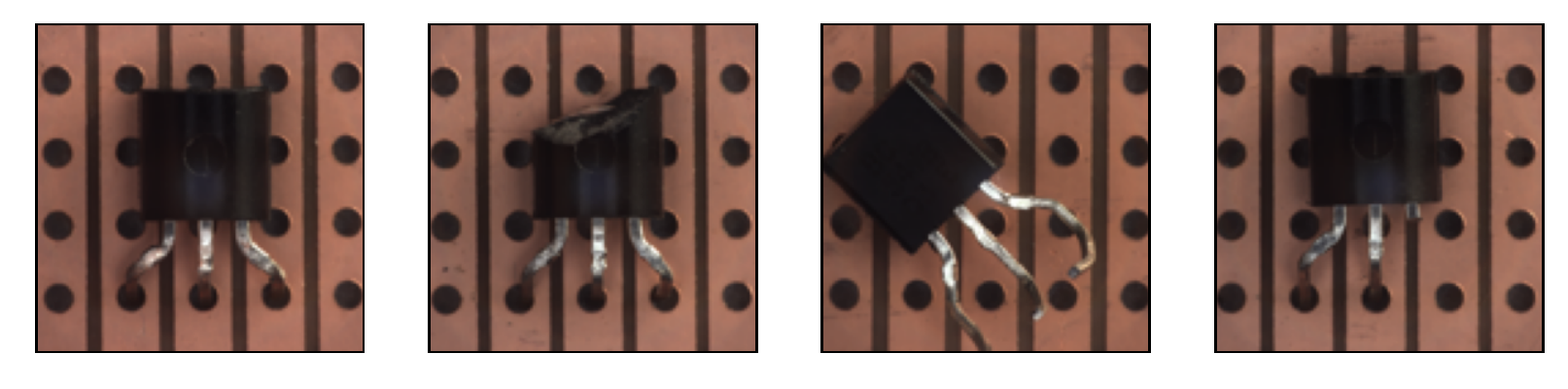}
	\caption{The first image from left is the defect-free transistor (normal class), while the rest three images present transistors with different defect types. Both normal and abnormal images have similar backgrounds.}
	\label{fig:example-transistor}
\end{figure}

Over the last few years, there have been already many previous studies in semi-supervised anomaly detection such as~\cite{sabokrou2018adversarially,schlachter2019one,zhou2017anomaly}. Although these methods have different architectures or learning objectives, one interesting observation is that they frequently consist of an autoencoder structure. This design is natural and intuitive because autoencoders are capable of learning the intrinsic structure of data, and this property can be used to detect abnormal samples during inference. However, many of these methods use autoencoders either to learn good latent representations fulfilling certain conditions or to minimize some reconstruction errors by using all features (e.g., all pixels of image data). This might lead to suboptimal anomaly detection performance, because not all information of the input data is necessary and useful. For example, as shown in Fig.~\ref{fig:example-transistor}, the background pixels of a transistor do not have to be well encoded or reconstructed in order to reject abnormal samples, because both abnormal and normal transistor images have similar or even the same background. Therefore, a technique that can automatically focus on the most discriminative parts of input data while assigning less or no attention to the less important parts is expected to be capable of enhancing the anomaly detection performance of all autoencoder-based approaches.

In this paper, a novel selection-and-weighting-based anomaly detection (SWAD) framework is proposed. In contrast to many previous studies, we carry out selection and weighting in a latent space learned by an autoencoder. In this way, the reconstructed normal data from the selected and weighted latent representations are expected to have significantly smaller reconstruction errors than those of unseen abnormal data. Correspondingly, the anomalies are expected to be more effectively rejected during inference. In summary, the major contributions of this work are:
\begin{itemize}
	\item A novel selection-and-weighting-based anomaly detection framework is proposed by introducing feature selection and weighting mechanism in latent space.
	\item SWAD is a generic framework that can be used as a wrapper for many existing autoencoder-based anomaly detection approaches.
	\item Systematic study on the key hyperparameters of SWAD was carried out on different datasets to show the superior performance of SWAD.
\end{itemize}
%\section{Related Work/Backgrounds}
%\label{sec:related-work}
%\subsection{Autoencoder-Based Anomaly Detection}
%Anomaly detection has been intensively studied in literature over the last decades~\cite{}. In some literature, anomaly detection can also refer to outlier detection~\cite{} or fault detection~\cite{}. In this paper, we stick to the term anomaly detection for consistency.
%
%\textbf{literature review on AAD}

\section{Autoencoder-Based Anomaly Detection}
\label{sec:aad}
In this section, we briefly review the basic idea of autoencoder-based anomaly detection methods. Generally, it is assumed that an autoencoder can well capture the hidden structure of the normal data, if the autoencoder is trained on these normal data only. After the training, the autoencoder can therefore well reconstruct the normal data, while it is expected to be incapable of correctly reconstructing abnormal samples. Consequently, by comparing the reconstruction errors between a test sample and its reconstruction, anomalies can be detected; i.e., normal samples typically have smaller reconstruction errors than those of abnormal samples. 

Formally, a training dataset is denoted as $X =[\bm{x}_1, \bm{x}_2, \dots, \bm{x}_N]^\top\in\mathbb{R}^{N\times D}$, where each sample $\bm{x}_i\in\mathbb{R}^D$ is from the normal class. An autoencoder is typically composed of an encoder $f_\text{enc}(\cdot; \bm{\Theta}_\text{enc})$ and a decoder $f_\text{dec}(\cdot; \bm{\Theta}_\text{dec})$ parameterized by $\bm{\Theta} = \{\bm{\Theta}_\text{enc}, \bm{\Theta}_\text{dec}\}$. The encoder first maps the input data $\bm{x}_i$ into a latent representation $\bm{z}_i = f_\text{enc}(\bm{x}_i)$ with $\bm{z}_i\in\mathbb{R}^L$, and $L$ is typically significantly smaller than the original dimension $D$. Subsequently, the decoder learns to map the latent representation back to the original space as $\hat{\bm{x}}_i = f_\text{dec}(\bm{z}_i)$. Accordingly, the overall learning objective of autoencoder is to minimize the reconstruction errors between the original input $\bm{x}_i$ and its reconstruction $\hat{\bm{x}}_i$ over the entire training dataset. One frequently used loss function is the mean squared error (MSE) loss defined as
\begin{equation}
	\label{eq:ae}
	\mathcal{L}_\text{AE} = \frac{1}{N}\sum_{i=1}^{N}||\bm{x}_i - \hat{\bm{x}}_i||^2_2.
\end{equation}
During inference, given a test sample denoted as $\bm{x}_\text{test}$, its reconstruction can be obtained as $\hat{\bm{x}}_\text{test} = f_\text{dec}(f_\text{enc}(\bm{x}_\text{test}))$. Thereby, the reconstruction error is calculated as $\varepsilon_\text{test} = ||\bm{x}_\text{test} - \hat{\bm{x}}_\text{test}||^2_2$. According to a predefined threshold $\varepsilon_0$, the test sample is detected as anomaly if $\varepsilon_\text{test}>\varepsilon_0$, whereas the test sample is affiliated with the normal class if $\varepsilon_\text{test}\leqslant\varepsilon_0$. It should be noted that the threshold is, in practice, either determined by a separate validation set with abnormal samples, or set by special requirements of certain use cases or practitioners. 
\section{Method}
\label{sec:method}
The key idea of the proposed method is to introduce feature selection to the learned latent representations, and we can thus identify the important (selected) and the less informative (non-selected) features in the latent space. Accordingly, the selected and non-selected features of the latent representations are separately weighted. Subsequently, the resulting weighted latent representations are fed into the trained decoder to reconstruct the inputs. By comparing the differences between the input data and reconstructions obtained by the weighted latent representations, we can therefore detect anomalies. In the following, we first present the proposed framework and then explain the implementation details of each individual module within the framework.

% %%%%%%%%%%%%%%%%%%%%%%%%%%%%%%%%%%%%%%%%%%%%%%%%%%%%%%%%%%%%%%%%%%%%%%%%%%%%%%
\subsection{SWAD Framework}
\label{subsec:swad-framework}
\begin{figure}[!ht]
	\centering
	\includegraphics[width=0.75\linewidth]{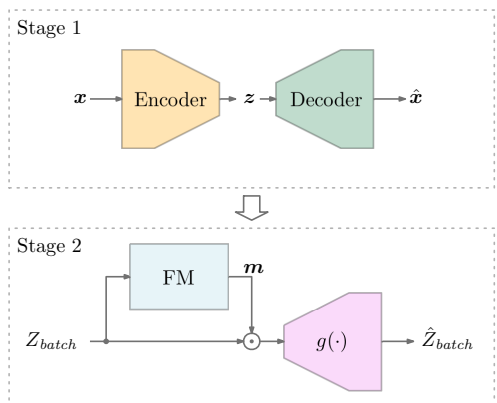}
	\caption{The structure of the proposed SWAD framework. }
	\label{fig:swad}
\end{figure}

The novel Selection-and-Weighting-based Anomaly Detection (SWAD) framework is composed of two major parts as shown in Fig.~\ref{fig:swad}: \emph{i}) an autoencoder as a backend model for learning the latent representations of normal data; \emph{ii}) a feature selection model for identifying the most informative features from the less important features in the learned latent space and for the subsequent weighting during inference. 
%The corresponding pseudo code for training is demonstrated in Algorithm~\ref{algo:training}.

\subsubsection{Training}
The training procedure is designed as a two-stage process. At the first stage, an autoencoder is trained on the given normal data, aiming to minimize a reconstruction loss. After training the autoencoder, we use the trained encoder to obtain the latent representations of the training data. At the second stage, the latent representations are used as new inputs to train a feature selection model. Consequently, the feature selection results can identify the more essential dimensions (\emph{selected} features) of the normal data in the latent space, and indicate the redundant or noisy dimensions (\emph{non-selected} features) that may mislead the anomaly detection. Therefore, by retaining the values of the selected features while down-weighting the non-selected features in latent space, it is expected to highlight the most discriminative characteristics of the normal data.

%caption for the figure above: The first stage is to train an autoencoder on the normal data. During the second stage, we train the FM-module~\cite{liao2020feature} on the learned latent representations to select the most informative features. Here, $Z_\mathit{batch} = [\bm{z}_1, \bm{z}_2, \dots, \bm{z}_{N_B}]^\top$ denotes the latent representations of a minibatch during each training iteration.

\subsubsection{Inference}
Two hyperparameters $k$ and $\tau$ are additionally introduced to carry out inference after training the SWAD framework. $k$ is the number of the most important features (i.e., the selected features) in the latent space, and $\tau\in(0,1)$ is the weighting factor for the non-selected features. To weight the latent representations, a fixed $\tau$ is multiplied to the non-selected features, while the $k$ selected features retain their values. In this case, it can be understood that the selected features are multiplied with $\tau=1$ to retain their values, while the non-selected features are down-weighted by $\tau < 1$. This is different from the ordinary feature selection, where non-selected features are typically removed or set to zeros. The proposed selection and weighting procedure can be therefore interpreted as a \emph{soft selection}. The soft selection is expected to be more practical for anomaly detection because we frequently have only limited knowledge about the anomalies. Then, the weighted latent representations are fed into the trained decoder $f_\text{dec}(\cdot)$ from the first stage to obtain reconstructions. Consequently, the reconstruction errors between the original inputs and the reconstructions are calculated as an indicator of the affiliation of the data. In a semi-supervised anomaly detection setup~\cite{chalapathy2019deep}, we use a separate validation set (consisting of a few abnormal samples) to determine the threshold. Accordingly, a test sample with a reconstruction error greater than the threshold is detected as an abnormal sample.

\subsection{Autoencoder}
\label{subsec:autoencder}
The autoencoder is the core backend model for the first training stage. Generally, the proposed SWAD framework has no special requirements on the architectures of autoencoders. Thereby, the formal structure of the autoencoder is the same as the one introduced in Section~\ref{sec:aad}, and the MSE loss defined in Eq.~\ref{eq:ae} can be used as the learning objective for SWAD. 

Nonetheless, we implemented a shallow autoencoder to demonstrate the effectiveness of the SWAD framework. In particular, the autoencoder consisted of three fully connected layers with 256, 128 and 256 neurons, respectively. A LeakyReLU layer with the rate of 0.2 was used as the activation layer after first and the last fully connected layer, while the activation function for the second hidden layer was $\mathit{sigmoid}(\cdot)$ to guarantee a bounded latent representation, which is more meaningful for training feature selection model at the second stage as suggested in~\cite{liao2020feature}.

% %%%%%%%%%%%%%%%%%%%%%%%%%%%%%%%%%%%%%%%%%%%%%%%%%%%%%%%%%%%%%%%%%%%%%%%%%%%%%%
\subsection{Selection}
\label{subsec:selection}
In order to identify the important features of the latent representations more effectively, we use the recently proposed feature mask (FM) method~\cite{liao2020feature} as the feature selection model for the second training stage. The original FM method consists of two major parts: an FM-module $f_\text{FM}(\cdot;\bm{\Theta}_\text{FM})$ to generate the feature mask and a learning network targeting classification or regression. In this work, we followed the architecture of FM-module in~\cite{liao2020feature}, while replaced the learning network by a decoder sharing the same architecture as the one in the autoencoder for the first training stage denoted as $g(\cdot;\bm{\Theta}_\text{g})$, where the parameters $\bm{\Theta}_\text{FM}$ and $\bm{\Theta}_\text{g}$ are learned during the second training stage. It should be noted that the learning network $g(\cdot)$ has the same architecture with the decoder $f_\text{dec}(\cdot)$ from the first training stage only for a simpler implementation. In general, these two networks are independent of each other. 

Specifically, the inputs of the FM-module are the learned latent representations $\bm{z}_i$ of the training data $\bm{x}_i$ from the first training stage. Subsequently, the feature mask $\bm{m}$ generated by FM-module is element-wisely multiplied with the latent representations and fed into the learning network $g(\cdot)$ to reconstruct the original training data. Correspondingly, during each training iteration, we aim to minimize
\begin{equation}
	\label{eq:selection}
	\mathcal{L}_\text{SE} = \frac{1}{N_B}\sum_{i=1}^{N_B}||\bm{x}_i - g\big(\bm{z}_i\odot f_\text{FM}(Z_\text{batch};\bm{\Theta}_\text{FM});\bm{\Theta}_\text{g}\big)||^2_2,
\end{equation}
where $Z_\text{batch} = [\bm{z}_1, \bm{z}_2, \dots, \bm{z}_{N_B}]^\top$ is a minibatch of learned latent representations of the training data during each iteration, and $N_B$ is the minibatch size. After the second training stage, the final feature mask $\bm{m}$ can be calculated by applying the learned $f_\text{FM}(\cdot)$ on the latent representations of the entire training data, where $\bm{m}$ is automatically bounded in the range of $[0, 1]$. Thereby, the $i$-th element of $\bm{m}$ indicates the importance of the corresponding latent feature. Accordingly, we can select the top $k$ features with the highest importance scores, whereas the left $L-k$ features are considered as non-selected (less important) features. It should be noted that, analogous to training autoencoder at the first stage, we use the MSE loss in Eq.~\ref{eq:selection} to demonstrate one possible implementation and other learning objectives for reconstructions can be used as well.

% %%%%%%%%%%%%%%%%%%%%%%%%%%%%%%%%%%%%%%%%%%%%%%%%%%%%%%%%%%%%%%%%%%%%%%%%%%%%%%
\subsection{Weighting and Detection}
\label{subsec:weighting}
Although the feature mask $\bm{m}$ generated by the FM-module can already identify the most informative latent features, as mentioned before, it can be hard for detecting anomalies if we trivially remove the non-selected features as in conventional feature selection. This is because the encoding procedure of the autoencoder takes into account the information of entire data, i.e. all original features. Therefore, even the non-selected features of the latent representations can still possess certain information about the normal class. Due to this concern, the latent representations are weighted based on the selection results. In particular, the weighting process is defined as
\begin{equation}
	\label{eq:weighting}
	\tilde{z}_{i,j}=\left\{
	\begin{array}{cl}
		z_{i, j} & \text{, if } z_{i, j} \in \text{selected $k$ features} \\
	    \tau\cdot z_{i, j} & \text{, otherwise}
	\end{array} 
	\right.
\end{equation}
where $z_{i, j}$ is the $j$-th element of the learned latent representation $\bm{z}_i$ of the sample $\bm{x}_i$. In this way, the resulting weighted latent representation is denoted as $\tilde{\bm{z}}_i = [\tilde{z}_{i,1}, \tilde{z}_{i,2}, \dots, \tilde{z}_{i,L}]^\top$. Consequently, for each test sample $\bm{x}_i$, the reconstruction error is calculated as:
\begin{equation}
	\label{eq:rec_error}
	\varepsilon_i = ||\bm{x}_i - f_\text{dec}(\tilde{\bm{z}}_i)||^2_2.
\end{equation}
Finally, as introduced in Section~\ref{sec:aad}, a test sample is rejected as anomaly if $\varepsilon_i > \varepsilon_0$. 
%It should be noted that the reconstruction error in Eq.~\ref{eq:rec_error} is one common measure. In practice, other metrics can also be used to calculate the reconstruction error such as structural similarity index measure (SSIM)~\cite{wang2004image}.

\section{Experiments}
\label{sec:exp}
This section evaluates the proposed novel SWAD framework for anomaly detection. Specifically, we aim to empirically show: \emph{i}) the SWAD framework can significantly improve the anomaly detection performance on both benchmark and real-world datasets; \emph{ii)} the SWAD framework has comparable or even better performance than state-of-the-art approaches; \emph{iii}) a systematic study of the newly introduced hyperparameters $k$ and $\tau$ of the SWAD framework.

% %%%%%%%%%%%%%%%%%%%%%%%%%%%%%%%%%%%%%%%%%%%%%%%%%%%%%%%%%%%%%%%%%%%%%%%%%%%%%%
\subsection{Setup}
\label{subsec:setup}
In this work, we followed the semi-supervised anomaly detection setup, meaning that only normal samples were used for training, while used a separate validation set (consisting of both normal and a few abnormal samples) for each experiment to stop the training when the model reached the best performance on the validation set. Specifically, we only report the results on the test set in the following. AUC~\cite{bradley1997use} was used as the metric to measure the anomaly detection performance as in the previous studies~\cite{ruff2018deep,schlachter2019deep}. In all experiments, we used the Adam optimizer~\cite{kingma2014adam} with a learning rate of $10^{-3}$. The batch size was 512 for the benchmark datasets and 256 for the real-world datasets. The key hyperparameters of the SWAD framework and the vanilla autoencoder were individually optimized on the validation set for each experiment. The implementation was based on TensorFlow~\cite{abadi2016tensorflow}. It should be noted that each experiment was repeated five times with different random seeds, and only the averaged results with their standard deviations are presented in the following subsections.

\subsubsection{Datasets}
In this study, the experiments were carried out on four datasets, including two image datasets MNIST~\cite{lecun1998gradient} and CIFAR-10~\cite{krizhevsky2009learning} which are frequently used to benchmark deep learning algorithms, and two challenging real-world datasets. Specifically, the two benchmark datasets have originally 10 classes. Therefore, in order to create an anomaly detection setup, data from one specific class are considered as normal, while the data affiliated with the left 9 classes were considered abnormal. The two real-world datasets are solar cell images~\cite{bartler2018automated} and MVTec AD~\cite{bergmann2019mvtec}. On both real-world datasets, normal samples are defect-free products, while the abnormal samples are products with different defects.

\subsubsection{Reference Methods}
The key reference method is a vanilla autoencoder (AE) with the same architecture as the one used in the SWAD framework. In addition, we also include two representative methods from literature to identify the anomaly detection ability of SWAD on benchmark datasets. One is the deep support vector domain description (DSVDD)~\cite{ruff2018deep} and the other is the one-class support vector machine (OCSVM)~\cite{scholkopf2001estimating}.

\begin{table}[ht]
	\renewcommand{\arraystretch}{2.2}
	\centering
	\caption{AUC on the MNIST dataset.}
	\label{tab:basic_exp}
	\resizebox{\columnwidth}{!}{
		\begin{tabular}{lccccc}
			\hline
			\bfseries & \bfseries DSVDD~\cite{ruff2018deep} & \bfseries OCSVM~\cite{scholkopf2001estimating} & \bfseries AE & \bfseries SWAD& \bfseries Gain\\
			\hline
			Digit 0 & \makecell[c]{0.980 \\ {\scriptsize($\pm$ 0.007)}}& \makecell[c]{0.982 \\ {\scriptsize($\pm$ 0.000)}}& \makecell[c]{0.988 \\ {\scriptsize($\pm$ 0.000)}}& \makecell[c]{\textbf{0.996} \\ {\scriptsize($\pm$ 0.001)}} & 0.8\%$\uparrow$\\ 
			Digit 1 & \makecell[c]{0.997 \\ {\scriptsize($\pm$ 0.001)}}& \makecell[c]{0.992 \\ {\scriptsize($\pm$ 0.000)}}& \makecell[c]{0.999 \\ {\scriptsize($\pm$ 0.001)}}& \makecell[c]{\textbf{1.000} \\ {\scriptsize($\pm$ 0.000)}} & 0.1\%$\uparrow$ \\
			Digit 2 & \makecell[c]{0.917 \\ {\scriptsize($\pm$ 0.008)}}& \makecell[c]{0.821 \\ {\scriptsize($\pm$ 0.000)}}& \makecell[c]{0.931 \\ {\scriptsize($\pm$ 0.003)}}& \makecell[c]{\textbf{0.937} \\ {\scriptsize($\pm$ 0.009)}} & 0.6\%$\uparrow$ \\ 
			Digit 3 & \makecell[c]{0.919 \\ {\scriptsize($\pm$ 0.015)}}& \makecell[c]{0.861 \\ {\scriptsize($\pm$ 0.000)}}& \makecell[c]{0.960 \\ {\scriptsize($\pm$ 0.002)}}& \makecell[c]{\textbf{0.971} \\ {\scriptsize($\pm$ 0.003)}} & 1.1\%$\uparrow$ \\
			Digit 4 & \makecell[c]{0.949 \\ {\scriptsize($\pm$ 0.008)}}& \makecell[c]{0.948 \\ {\scriptsize($\pm$ 0.000)}}& \makecell[c]{0.945 \\ {\scriptsize($\pm$ 0.002)}}& \makecell[c]{\textbf{0.969} \\ {\scriptsize($\pm$ 0.004)}} & 2.5\%$\uparrow$ \\ 
			Digit 5 & \makecell[c]{0.885 \\ {\scriptsize($\pm$ 0.009)}}& \makecell[c]{0.774 \\ {\scriptsize($\pm$ 0.000)}}& \makecell[c]{0.975 \\ {\scriptsize($\pm$ 0.000)}}& \makecell[c]{\textbf{0.993} \\ {\scriptsize($\pm$ 0.000)}} & 1.8\%$\uparrow$ \\
			Digit 6 & \makecell[c]{0.983 \\ {\scriptsize($\pm$ 0.005)}}& \makecell[c]{0.948 \\ {\scriptsize($\pm$ 0.000)}}& \makecell[c]{0.996 \\ {\scriptsize($\pm$ 0.001)}}& \makecell[c]{\textbf{0.997} \\ {\scriptsize($\pm$ 0.001)}} & 0.1\%$\uparrow$ \\ 
			Digit 7 & \makecell[c]{0.946 \\ {\scriptsize($\pm$ 0.009)}}& \makecell[c]{0.934 \\ {\scriptsize($\pm$ 0.000)}}& \makecell[c]{0.971 \\ {\scriptsize($\pm$ 0.005)}}& \makecell[c]{\textbf{0.987} \\ {\scriptsize($\pm$ 0.002)}} & 1.6\%$\uparrow$ \\
			Digit 8 & \makecell[c]{\textbf{0.939} \\ {\scriptsize($\pm$ 0.016)}}& \makecell[c]{0.902 \\ {\scriptsize($\pm$ 0.000)}}& \makecell[c]{0.912 \\ {\scriptsize($\pm$ 0.006)}}& \makecell[c]{0.936 \\ {\scriptsize($\pm$ 0.023)}} & 2.6\%$\uparrow$ \\ 
			Digit 9 & \makecell[c]{0.965 \\ {\scriptsize($\pm$ 0.003)}}& \makecell[c]{0.928 \\ {\scriptsize($\pm$ 0.000)}}& \makecell[c]{0.963 \\ {\scriptsize($\pm$ 0.008)}}& \makecell[c]{\textbf{0.976} \\ {\scriptsize($\pm$ 0.002)}} & 1.3\%$\uparrow$ \\
			\hdashline
			Average & \makecell[c]{0.948 \\ {\scriptsize($\pm$ 0.008)}}& \makecell[c]{0.909 \\ {\scriptsize($\pm$ 0.000)}}& \makecell[c]{0.964 \\ {\scriptsize($\pm$ 0.003)}}& \makecell[c]{\textbf{0.976} \\ {\scriptsize($\pm$ 0.005)}} & 1.3\%$\uparrow$ \\
			\hline
	\end{tabular}}
\end{table}

\begin{table}[ht]
	\renewcommand{\arraystretch}{2.2}
	\centering
	\caption{AUC on the CIFAR-10 dataset.}
	\label{tab:basic_exp_cifar10}
	\resizebox{\columnwidth}{!}{
		\begin{tabular}{lccccc}
			\hline
			\bfseries & \bfseries DSVDD~\cite{ruff2018deep} & \bfseries OCSVM~\cite{scholkopf2001estimating} & \bfseries AE & \bfseries SWAD& \bfseries Gain\\
			\hline
			Airplane & \makecell[c]{0.617 \\ {\scriptsize($\pm$ 0.041)}}& \makecell[c]{0.619 \\ {\scriptsize($\pm$ 0.000)}}& \makecell[c]{0.681 \\ {\scriptsize($\pm$ 0.001)}}& \makecell[c]{\textbf{0.733} \\ {\scriptsize($\pm$ 0.005)}} & 7.6\%$\uparrow$ \\
			Automobile & \makecell[c]{\textbf{0.659} \\ {\scriptsize($\pm$ 0.021)}}& \makecell[c]{0.385 \\ {\scriptsize($\pm$ 0.000)}}& \makecell[c]{0.494 \\ {\scriptsize($\pm$ 0.010)}}& \makecell[c]{0.551 \\ {\scriptsize($\pm$ 0.013)}} & 11.5\%$\uparrow$ \\
			Bird & \makecell[c]{0.508 \\ {\scriptsize($\pm$ 0.008)}}& \makecell[c]{0.606 \\ {\scriptsize($\pm$ 0.000)}}& \makecell[c]{0.669 \\ {\scriptsize($\pm$ 0.001)}}& \makecell[c]{\textbf{0.677} \\ {\scriptsize($\pm$ 0.005)}} & 1.2\%$\uparrow$ \\
			Cat & \makecell[c]{\textbf{0.591} \\ {\scriptsize($\pm$ 0.014)}}& \makecell[c]{0.494 \\ {\scriptsize($\pm$ 0.034)}}& \makecell[c]{0.581 \\ {\scriptsize($\pm$ 0.004)}}& \makecell[c]{0.584 \\ {\scriptsize($\pm$ 0.004)}} & 3.3\%$\uparrow$ \\
			Deer & \makecell[c]{0.609 \\ {\scriptsize($\pm$ 0.011)}}& \makecell[c]{0.713 \\ {\scriptsize($\pm$ 0.000)}}& \makecell[c]{0.707 \\ {\scriptsize($\pm$ 0.000)}}& \makecell[c]{\textbf{0.730}\\ {\scriptsize($\pm$ 0.002)}} & 1.5\%$\uparrow$ \\
			Dog & \makecell[c]{\textbf{0.657} \\ {\scriptsize($\pm$ 0.025)}}& \makecell[c]{0.520 \\ {\scriptsize($\pm$ 0.011)}}& \makecell[c]{0.607 \\ {\scriptsize($\pm$ 0.002)}}& \makecell[c]{0.616 \\ {\scriptsize($\pm$ 0.010)}} & 3.3\%$\uparrow$ \\
			Frog & \makecell[c]{0.677 \\ {\scriptsize($\pm$ 0.026)}}& \makecell[c]{0.638 \\ {\scriptsize($\pm$ 0.000)}}& \makecell[c]{0.668 \\ {\scriptsize($\pm$ 0.005)}}& \makecell[c]{\textbf{0.679} \\ {\scriptsize($\pm$ 0.001)}} & 1.6\%$\uparrow$ \\
			Horse & \makecell[c]{\textbf{0.673} \\ {\scriptsize($\pm$ 0.009)}}& \makecell[c]{0.482 \\ {\scriptsize($\pm$ 0.000)}}& \makecell[c]{0.493 \\ {\scriptsize($\pm$ 0.003)}}& \makecell[c]{0.501 \\ {\scriptsize($\pm$ 0.028)}} & 1.6\%$\uparrow$ \\
			Ship & \makecell[c]{\textbf{0.759} \\ {\scriptsize($\pm$ 0.012)}}& \makecell[c]{0.637 \\ {\scriptsize($\pm$ 0.000)}}& \makecell[c]{0.697 \\ {\scriptsize($\pm$ 0.001)}}& \makecell[c]{0.723 \\ {\scriptsize($\pm$ 0.007)}} & 3.7\%$\uparrow$ \\
			Truck & \makecell[c]{\textbf{0.731} \\ {\scriptsize($\pm$ 0.012)}}& \makecell[c]{0.488 \\ {\scriptsize($\pm$ 0.000)}}& \makecell[c]{0.425 \\ {\scriptsize($\pm$ 0.008)}}& \makecell[c]{0.568 \\ {\scriptsize($\pm$ 0.014)}} & 33.6\%$\uparrow$ \\
			\hdashline
			Average & \makecell[c]{\textbf{0.648} \\ {\scriptsize($\pm$ 0.018)}}& \makecell[c]{0.558 \\ {\scriptsize($\pm$ 0.000)}}& \makecell[c]{0.602 \\ {\scriptsize($\pm$ 0.004)}}& \makecell[c]{0.636 \\ {\scriptsize($\pm$ 0.009)}} & 6.9\%$\uparrow$ \\
			\hline
	\end{tabular}}
\end{table}

% %%%%%%%%%%%%%%%%%%%%%%%%%%%%%%%%%%%%%%%%%%%%%%%%%%%%%%%%%%%%%%%%%%%%%%%%%%%%%%
\subsection{Experiments on Benchmark Datasets}
\label{subsec:basic-exp}
The experiments conducted on benchmark datasets aim to demonstrate the effectiveness of the proposed framework and provide a preliminary comparison between SWAD and existing state-of-the-art approaches. TABLE~\ref{tab:basic_exp} and TABLE~\ref{tab:basic_exp_cifar10} show the entire experimental results on MNIST and CIFAR-10, respectively. Generally, the SWAD framework notably improves the anomaly detection performance of  the autoencoder. For example, on CIFAR-10, SWAD achieved about 6.9\% better performance than autoencoder in average with up to 33\% improvement in extreme cases (class truck as normal class). On MNIST, the improvement brought by the new framework is slightly less notable, namely about 1.3\% in average. This is reasonable because the vanilla autoencoder already performed well on the MNIST dataset. 

The comparison between SWAD and the vanilla autoencoder has empirically shown the benefit of the novel selection-and-weighting mechanism. This matches our expectation because the vanilla autoencoder trivially encodes all original features of the training data, while some information (e.g.,  background pixels or noise) can mislead the anomaly detection. By introducing the selection-and-weighting in the learned latent space, we could further identify the most important latent dimensions and therefore achieved notable improvement.

In addition, the SWAD framework also achieved comparable or even better performance in comparison with the two reference methods on the benchmark datasets. On MNIST, SWAD outperformed the reference methods in nine of ten cases. On the more challenging CIFAR-10 dataset, the proposed framework still had comparable performance with DSVDD and significantly outperformed OCSVM. It should be noted that we only used shallow fully-connected networks on the benchmark datasets to illustrate the effectiveness, so further performance enhancement can be expected when more powerful backbone networks are used such as convolutional neural networks.

% %%%%%%%%%%%%%%%%%%%%%%%%%%%%%%%%%%%%%%%%%%%%%%%%%%%%%%%%%%%%%%%%%%%%%%%%%%%%%%
\subsection{Case Study: Solar Cells}
\label{sec:solar}
\begin{figure}[!htb]
	\centering
	\includegraphics[width=.99\linewidth]{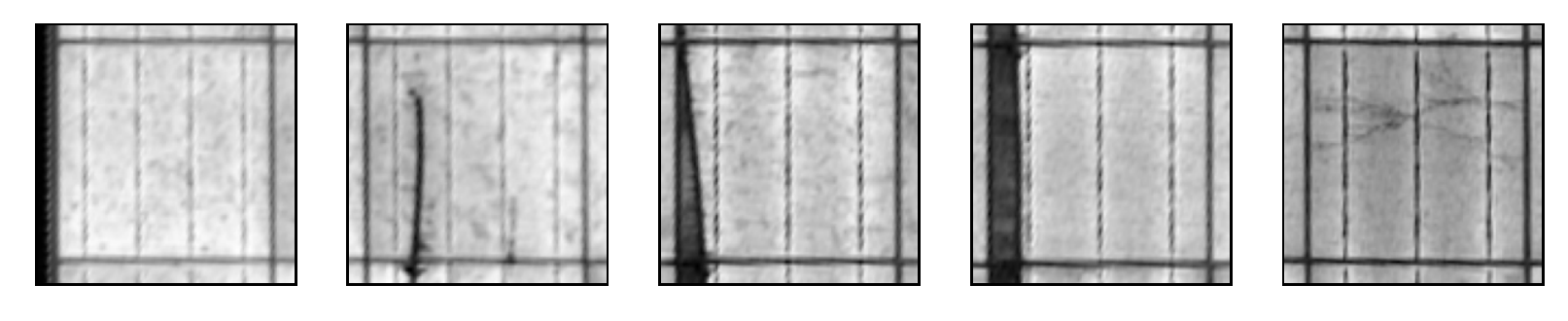}
	\caption{The first image from left is the normal solar cell, while the rest four images represent different defects.}
	\label{fig:example-solar}
\end{figure}

To automatically detect or identify defects on solar cells is an important use case for anomaly detection. In this experiment, we compared SWAD and a vanilla autoencoder on the solar cell image dataset~\cite{bartler2018automated}. The solar cell images were extracted from large-scale electroluminescence images. In total, 85513 normal solar cell images were used for training. Both validation and test sets consisted of 4737 normal and 189 abnormal images. Exemplary normal and abnormal solar cell images are shown in Fig.~\ref{fig:example-solar}. 

The solar cell images were more complex and had sizes of 120$\times$120, so the backbone networks in SWAD were replaced by convolutional autoencoders. In particular, the encoder consisted of 5 convolutional layers with 4, 8, 16, 32, 128 filters, respectively. The first four convolutional layers had filters with the size of 3$\times$3 using a stride of 2 with the same padding, while the last convolutional layer used a 5$\times$5 filter. Batch-normalization and Leaky-ReLU layers were used after each convolutional layer. The decoder used a mirrored structure with transposed convolutional layers. The learning network $g(\cdot)$ during the second training stage shared the same architecture with the decoder. As shown in TABLE~\ref{tab:solar}, the proposed SWAD framework improved the anomaly detection of an autoencoder moderately with respect to AUC.  

%The limited performance gain might be mainly due to the low resolution of the resized solar cell images: Many defects might be invisible and removed by the resizing procedure. We argue that training a larger network with the original resolution can result in more significant improvement by using our framework.

%\begin{table}[ht]
%	\renewcommand{\arraystretch}{2.2}
%	\centering
%	\caption{AUC on the solar cell dataset.}
%	\label{tab:solar}
%	%	\resizebox{\columnwidth}{!}{
%	\begin{tabular}{lcc}
%		\hline
%		\bfseries & \bfseries AE & \bfseries SWAD (ours)\\
%		\hline
%		Solar cells & \makecell[c]{0.846 \\ {\scriptsize($\pm$ 0.000)}}& \makecell[c]{0.854 \\ {\scriptsize($\pm$ 0.000)}} \\ 
%		\hline
%	\end{tabular}%}
%\end{table}

\begin{table}[ht]
	\renewcommand{\arraystretch}{2.2}
	\centering
	\caption{AUC on the solar cell dataset.}
	\label{tab:solar}
	%	\resizebox{\columnwidth}{!}{
	\begin{tabular}{lccc}
		\hline
		\bfseries & \bfseries AE & \bfseries SWAD (ours) & \bfseries Gain \\
		\hline
		Solar cells & \makecell[c]{0.744 \\ {\scriptsize($\pm$ 0.001)}}& \makecell[c]{0.754 \\ {\scriptsize($\pm$ 0.002)}} & 1.3\%$\uparrow$\\ 
		\hline
	\end{tabular}%}
\end{table}

\subsection{Case Study: MVTec}
\label{sec:mvtec}
\begin{figure}[!htb]
	\centering
	\includegraphics[width=.9\linewidth]{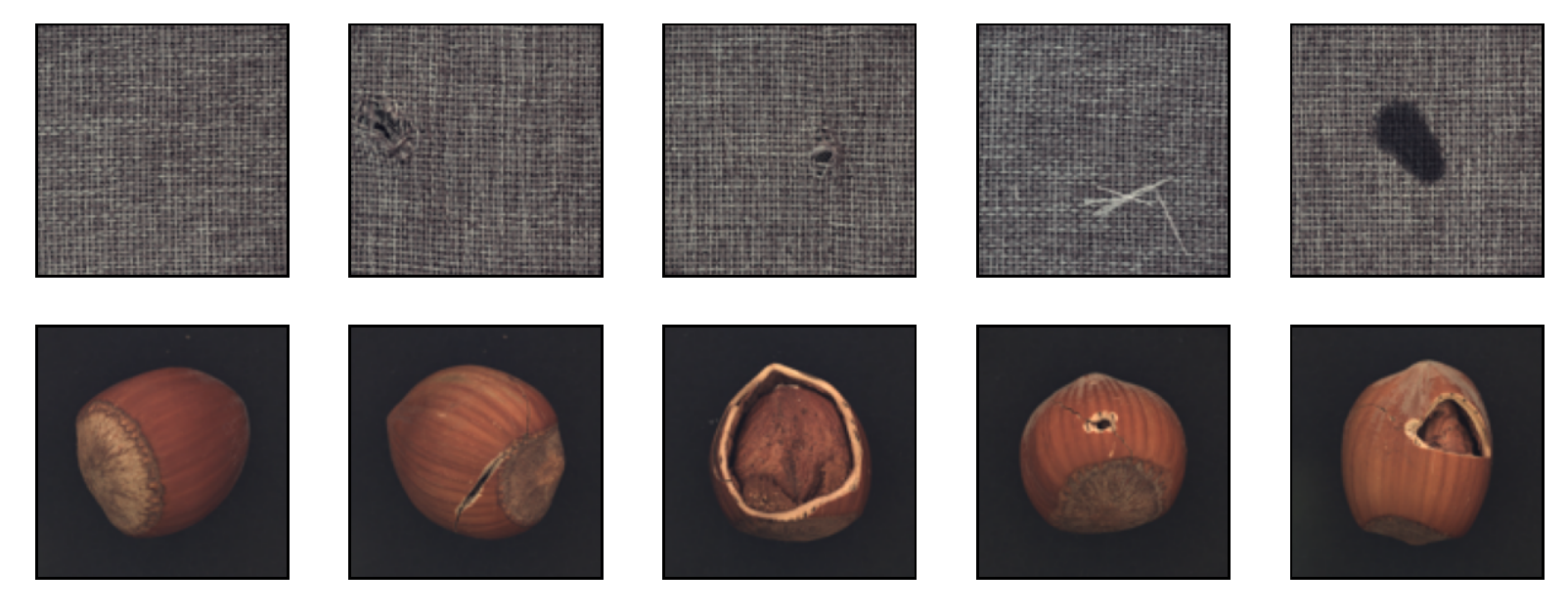}
	\caption{Exemplary MVTec images. The first column from the left shows the normal images, while the rest columns illustrate the abnormal samples.}
	\label{fig:example-mvtec}
\end{figure}

%152.2
%16.7
%27.9
%26.8
%2.9
%1.7
%19.5
%16.3
%4.0
%21.5
%8.6
%90.3
%1.1
%15.2
%3.6
%averages: 27.2
MVTec~\cite{bergmann2019mvtec} is a recently published real-world dataset for evaluating anomaly detection algorithms as shown in Fig.~\ref{fig:example-mvtec}. The original MVTec dataset consists of images with high resolution (1024$\times$1024) for 15 categories. In our experiment, the images were resized into 120$\times$120 to spare training resources. In addition, as suggested in~\cite{bergmann2019mvtec}, we used data augmentation to increase the training data size to 4000 for each category. Both validation and test sets had tens of normal and abnormal images (varying from category to category). Specifically, for each category, defect-free products were considered to be normal, while the products affiliated with the same category but with defects were denoted as abnormal samples. Moreover, SWAD was implemented as the same as for solar cell images.

\begin{figure}[!htb]
	\centering
	\includegraphics[width=1\linewidth]{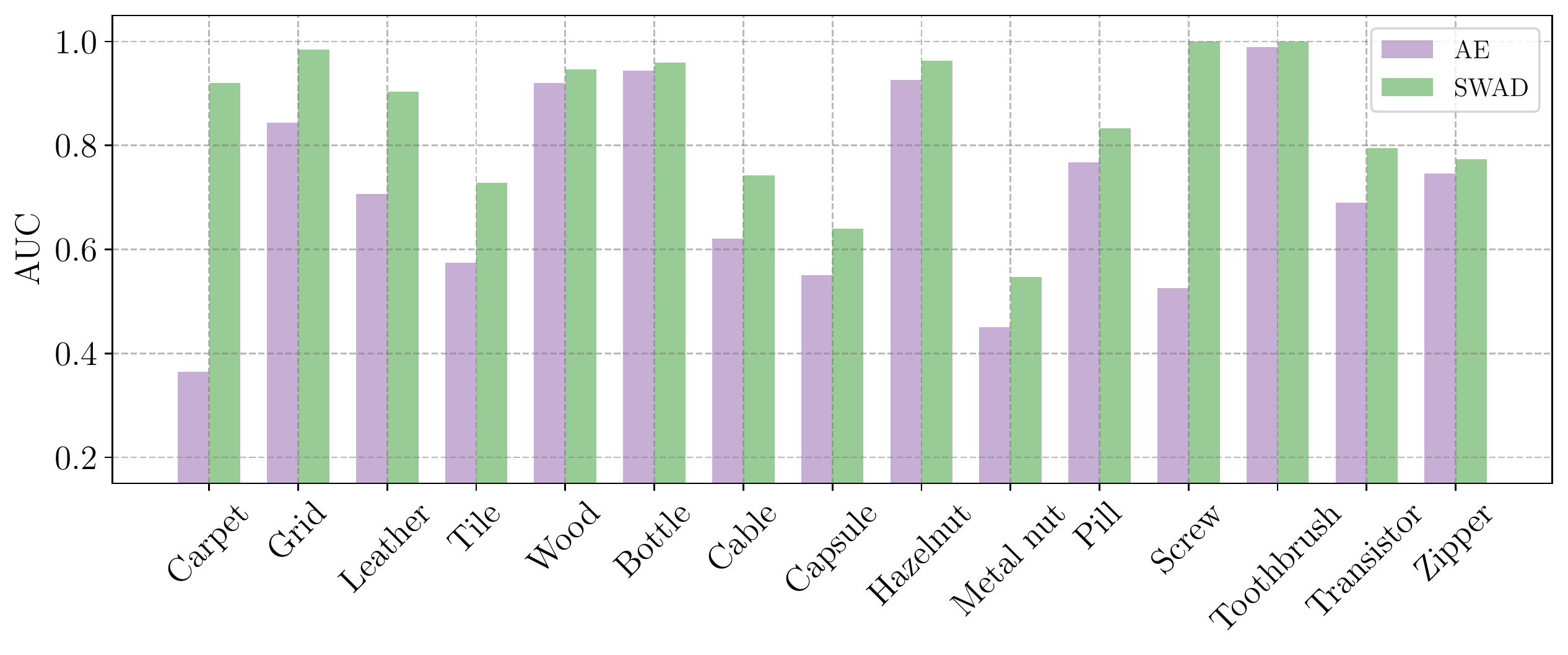}
	\caption{AUC on the MVTec dataset.}
	\label{fig:mvtec_res}
\end{figure}

Fig.~\ref{fig:mvtec_res} illustrates the overall AUC on the test sets for different categories of MVTec. The SWAD framework outperformed the vanilla autoencoder for all 15 categories with more than 27\% in average. In 10 of 15 cases, SWAD significantly improved the performance of a vanilla autoencoder with more than 5\%, where the largest improvement was about 152\% for the category \emph{carpet}. It should be noted that the SWAD framework only had marginally improvement with the categories \emph{wood}, \emph{bottle} and \emph{toothbrush}. One plausible reason might be the defect structure in these categories were too small and difficult to identify based on the current metric (MSE).

\subsection{Insight into SWAD}
\label{subsec:ablation}

%\begin{itemize}
%	\item rec errors distribution AE and SWAD:
%	\item misclassified samples
%	\item reconstruction comparisons
%\end{itemize}

\subsubsection{Hyperparameters $k$ and $\tau$}
\begin{figure}
	\centering
	\begin{subfigure}[b]{0.235\textwidth}
		\centering
		\includegraphics[width=\textwidth]{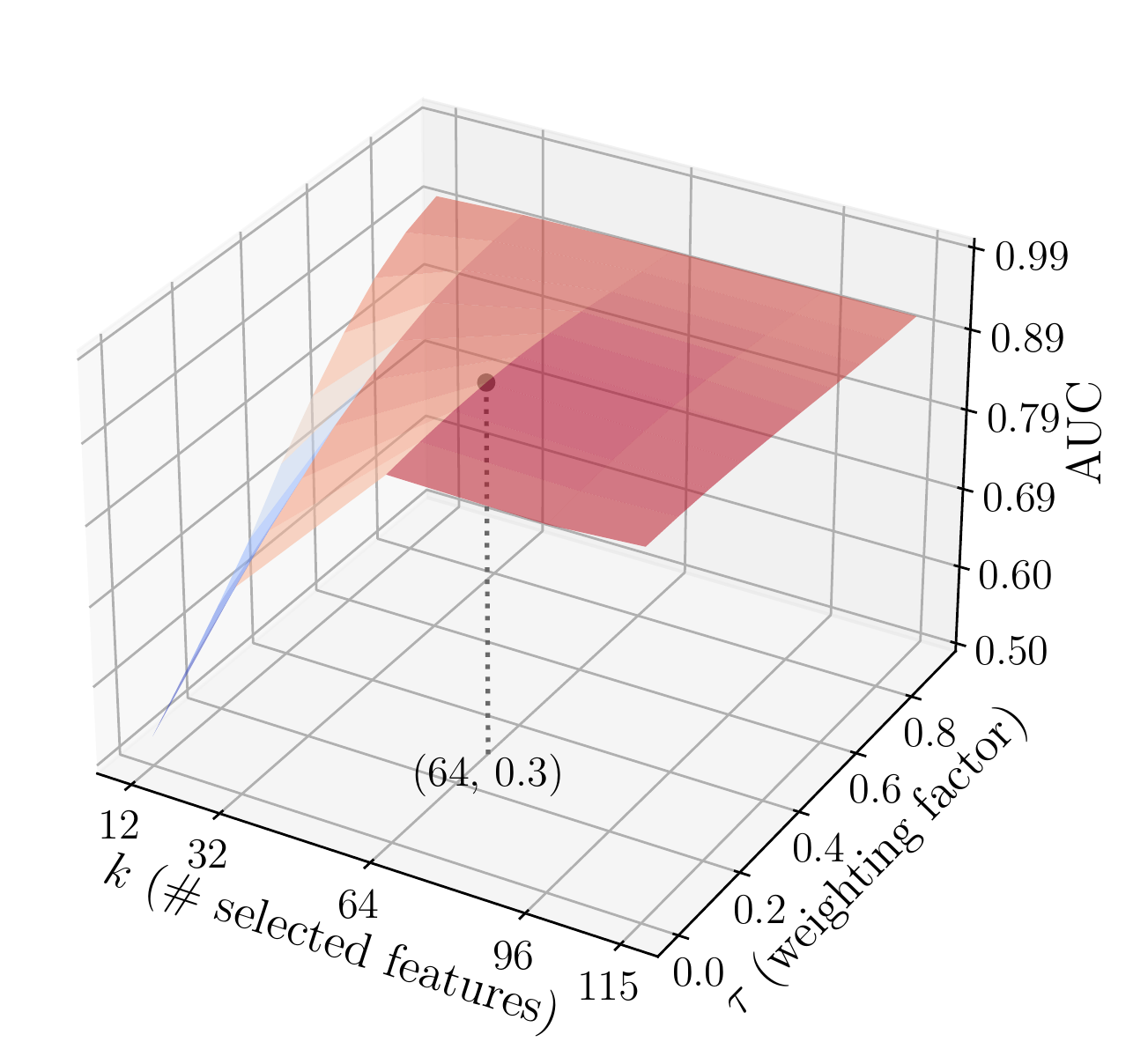}
		\caption{{\small MNIST (digit 8)}}    
%		\label{fig:}
	\end{subfigure}
	\hfill
	\begin{subfigure}[b]{0.235\textwidth}  
		\centering 
		\includegraphics[width=\textwidth]{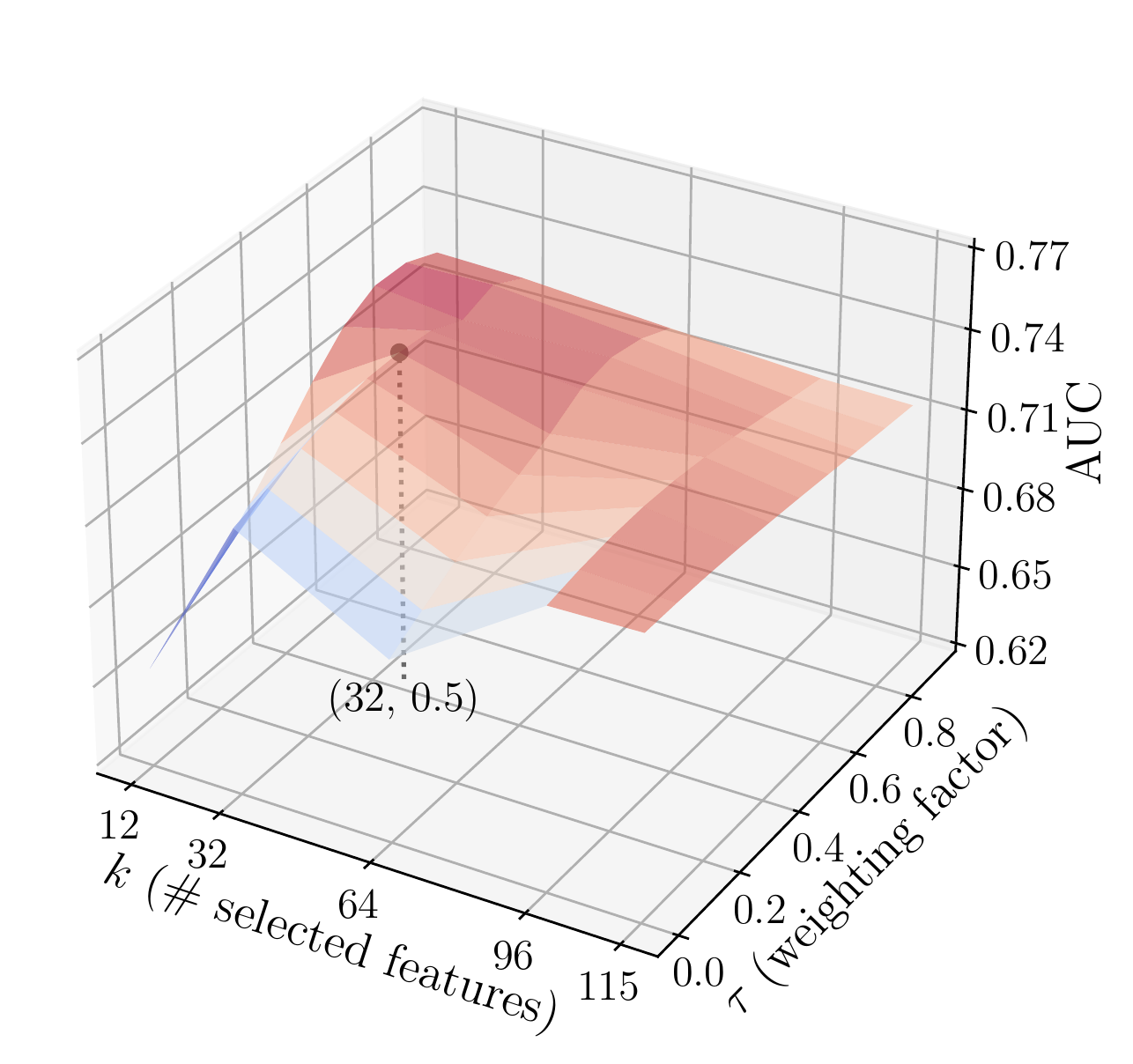}
		\caption{{\small CIFAR-10 (deer)}}
	\end{subfigure}
%	\vskip\baselineskip
	\begin{subfigure}[b]{0.235\textwidth}   
		\centering 
		\includegraphics[width=\textwidth]{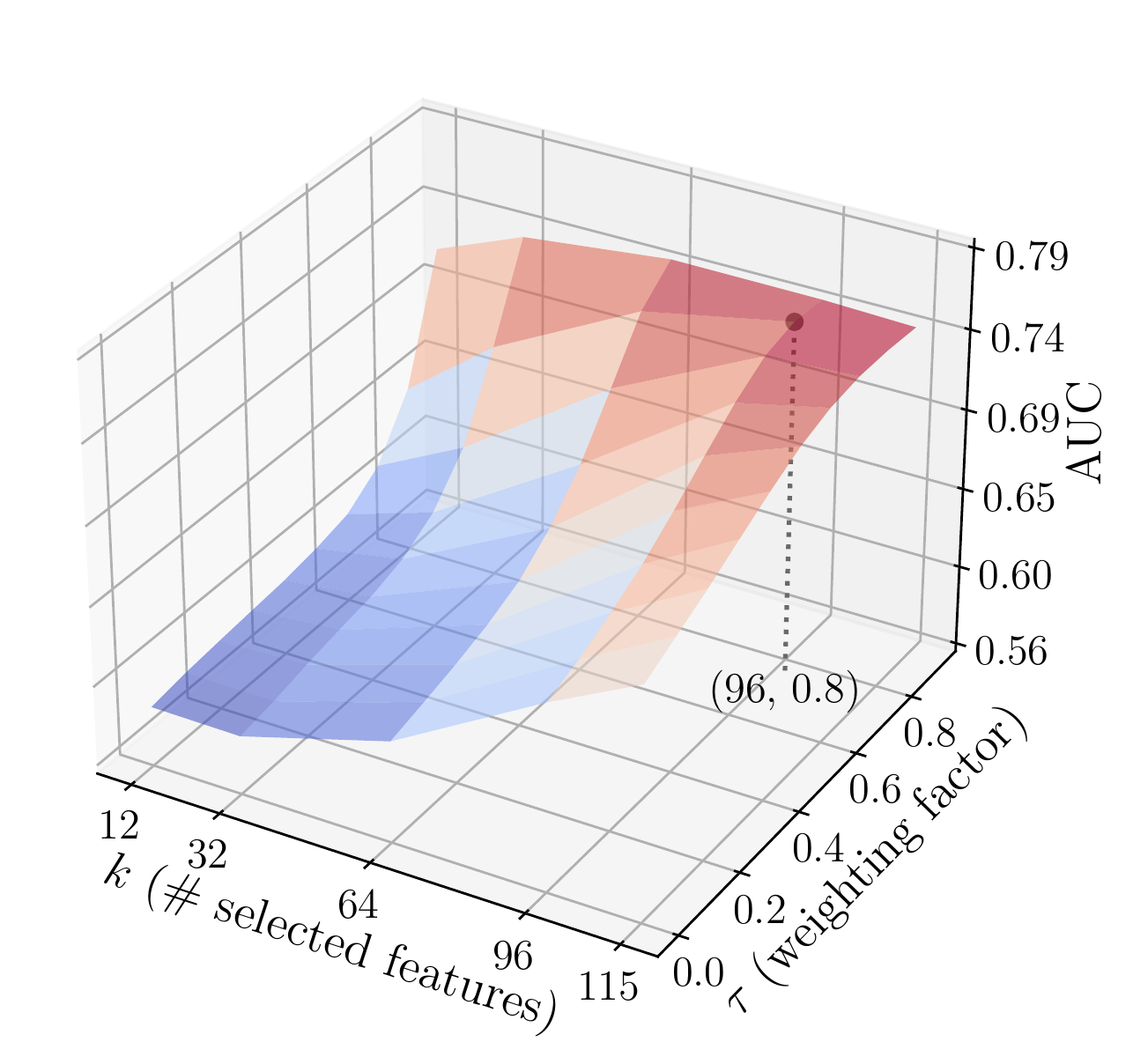}
		\caption{{\small Solar cells}}    
	\end{subfigure}
	\hfill
	\begin{subfigure}[b]{0.235\textwidth}   
		\centering 
		\includegraphics[width=\textwidth]{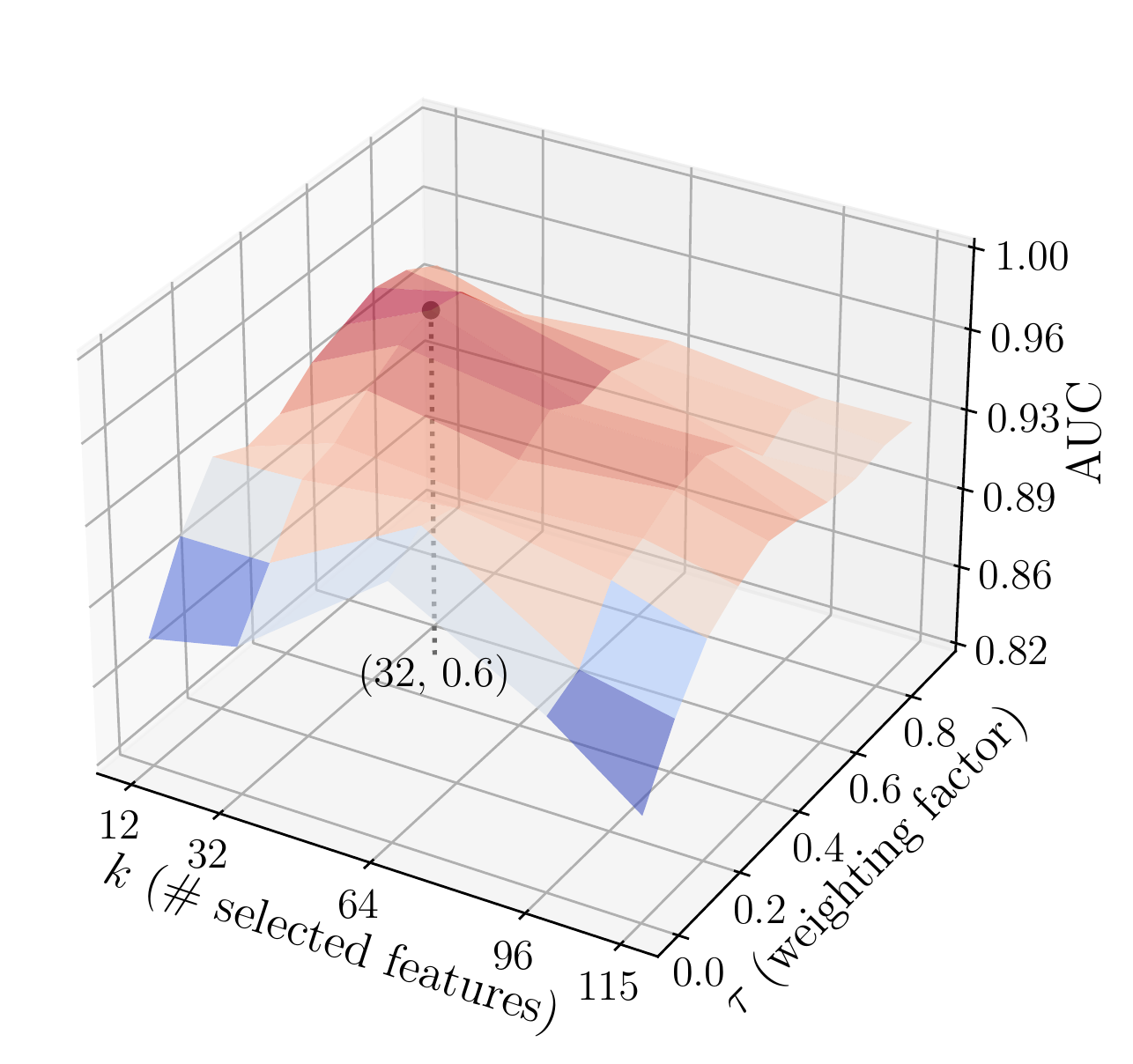}
		\caption{{\small MVTec (hazelnut)}}    
	\end{subfigure}
	\caption{AUC with respect to the two key hyperparameters $k$ and $\tau$ on four exemplary experiments. The highest AUC is marked using a black point with dotted lines.} 
	\label{fig:hyper_surface}
\end{figure}

Fig.~\ref{fig:hyper_surface} shows the exemplary performance overview for the four datasets. In general, the benchmark datasets are less challenging than the real-world datasets. Therefore, the resulting visualization of benchmark datasets is more homogeneous. On the contrary, the surface of the two real-world datasets is notably more bumpy. Furthermore, large $\tau$ cannot always guarantee the optimal anomaly detection performance. This matches our expectation that not all learned latent features can effectively contribute to detecting anomalies. In this case, the SWAD framework is essential to approach the optimal performance. 

\subsubsection{Study of Latent Dimensions}
\begin{figure}[!htb]
	\centering
	\includegraphics[width=.85\linewidth]{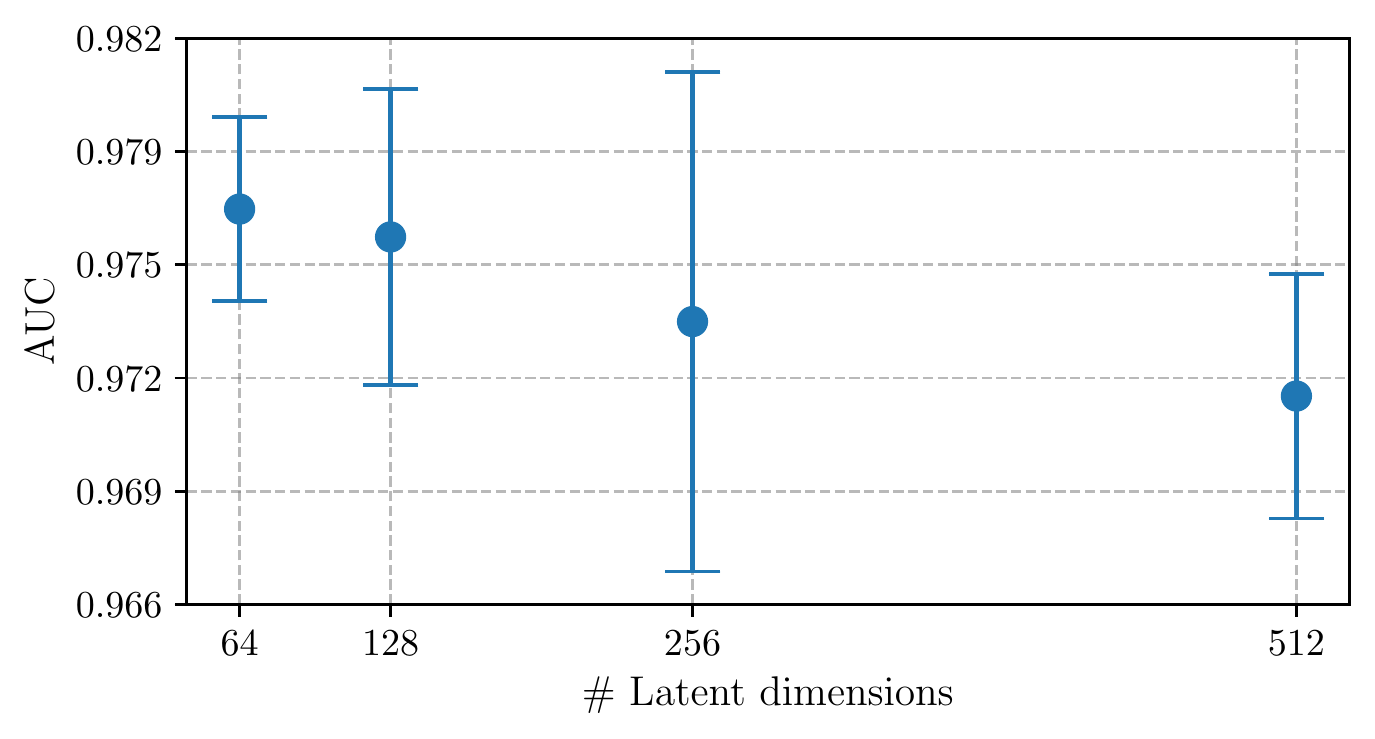}
	\caption{Averaged AUC on MNIST under different latent dimensions of the autoencoder at the first training stage.}
	\label{fig:mnist_lat}
\end{figure}

\begin{figure}[!htb]
	\centering
	\includegraphics[width=.85\linewidth]{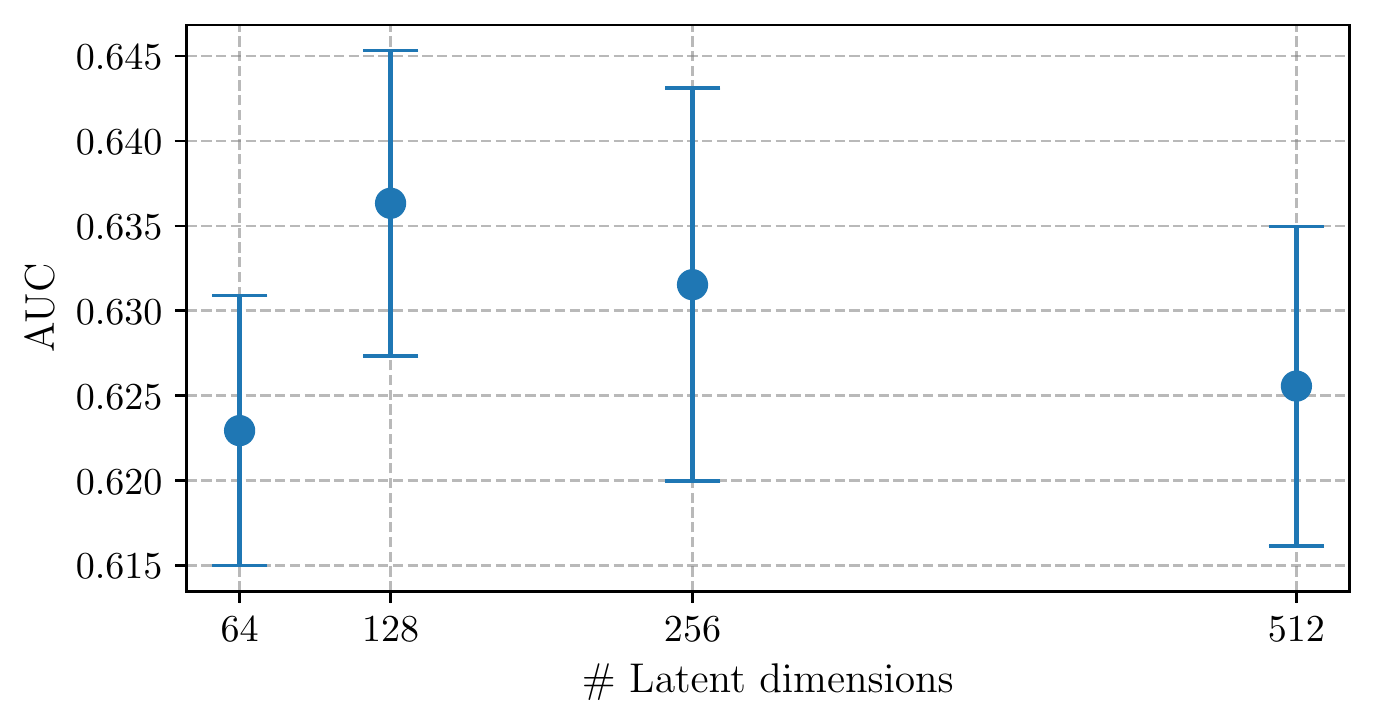}
	\caption{Averaged AUC on CIFAR-10 under different latent dimensions of the autoencoder at the first training stage.}
	\label{fig:cifar_lat}
\end{figure}

The proposed SWAD framework deals with the latent representation learned by an autoencoder. Accordingly, the latent dimensions can have impact on the final performance. We carried out experiments on the benchmark datasets with four different latent dimensions: 64, 128, 256 and 512. Fig.~\ref{fig:mnist_lat} and Fig.~\ref{fig:cifar_lat} present the results averaged over the ten classes for each dataset. $k$ and $\tau$ were individually optimized for each normal class. Generally, latent dimensions have limited influence on the anomaly detection performance. On both datasets, it can be observed that the averaged AUC decreases with large latent dimensions. One reason might be that too large latent dimensions cannot well encode the input data and the subsequent selection and weighting can thus have limited contribution. On the other hand, for the more challenging dataset CIFAR-10, too few latent dimensions cannot perform well either. This suggests that some information of the normal data is not well preserved during the encoding due to the small latent dimension. Nonetheless, the proposed framework can still well work in a wide range of latent dimensions.

% %%%%%%%%%%%%%%%%%%%%%%%%%%%%%%%%%%%%%%%%%%%%%%%%%%%%%%%%%%%%%%%%%%%%%%%%%%%%%%
\subsection{Discussion}
\label{subsec:discussion}
In the proposed SWAD framework, the selection-and-weighting is carried out in the learned latent space. Indeed, another intuitive alternative is to directly select and weight the features in the original data space, i.e., the pixel space for the image data. However, using pixels as features are less robust and more sensitive to the noise in the original images. Nonetheless, we argue that selection-and-weighting in the original data space can work if the data are structured. In addition, in this work, we used a fixed $\tau$ for a given dataset which was determined by a separate validation set. This design is intuitive and convenient for the implementation. Thereby, one natural future research is to include $\tau$ as a trainable parameter of the SWAD framework and the resulting framework is expected to be easier to use due to less hyperparameters.

As demonstrated in the previous subsections, SWAD has shown its ability of enhancing anomaly detection performance of autoencoder-based approaches. Nonetheless, we argue that our framework can also improve the representation-learning-based anomaly detection methods such as~\cite{ruff2018deep,chalapathy2018anomaly}. Although these approaches do not utilize autoencoders, they finally output a latent representation for each sample and calculate certain distances based on the learned representation to determine the affiliation of the sample. Correspondingly, these methods often share the similar drawback with the autoencoder-based approaches: All original input dimensions or features are equally considered and transformed into a compact latent representation, so noise or redundant information can be also encoded and thus mislead the anomaly detection process. To apply the SWAD framework to representation-learning-based methods, we can simply replace the autoencoder by the targeting methods at the first training stage. The general pipeline stays the same. In this way, the final distance calculation can be carried out on the selected-and-weighted representations and better performance can be expected. This is also one of the on-going work of the authors.

%future research directions
%\begin{itemize}
%	\item individual weighting factor
%	\item stochastic weighting factor
%	\item end-to-end model
%\end{itemize}
\section{Conclusion}
This paper proposes a novel selection-and-weighting-based anomaly detection framework, namely SWAD. In contrast to existing autoencoder-based approaches, SWAD aims to select the most informative features of latent representations and to down-weight the non-selected features for better anomaly detection performance. Experiments on both benchmark and real-world datasets have shown its effectiveness. Furthermore, the study of the additional hyperparameters of SWAD is also practical for users. The future work may include applying the SWAD framework to other autoencoder-based anomaly detection approaches. Another future work might introduce adaptive weighting for individual latent dimensions for further enhancement. Finally, other learning objectives such as adversarial training can be considered to replace the canonical MES loss for better performance.

\section*{Acknowledgment} This research was supported by Advantest as part of the Graduate School ``Intelligent Methods for Test and Reliability'' (GS-IMTR) at the University of Stuttgart. We would also like to show our gratitude to Solarzentrum Stuttgart GmbH for sharing their datasets.

\bibliographystyle{unsrt}
\bibliography{./Refs.bib}

\end{document}